\documentclass{article}

\usepackage{arxiv}
\usepackage[utf8]{inputenc} % allow utf-8 input
\usepackage[T1]{fontenc}    % use 8-bit T1 fonts
\usepackage{hyperref}       % hyperlinks
\usepackage{url}            % simple URL typesetting
\usepackage{booktabs}       % professional-quality tables
\usepackage{amsfonts}       % blackboard math symbols
\usepackage{amsmath}
\usepackage{amssymb}
\usepackage{nicefrac}       % compact symbols for 1/2, etc.
\usepackage{microtype}      % microtypography
\usepackage{cleveref}       % smart cross-referencing
\usepackage{lipsum}         % Can be removed after putting your text content
\usepackage{graphicx}
\usepackage[square,numbers]{natbib}
\usepackage{doi}
\usepackage{comment}
\usepackage{multirow}
\usepackage[export]{adjustbox}
\usepackage{bbm}
\usepackage{subcaption}
\usepackage[misc]{ifsym}
\usepackage{color}
\usepackage{pifont}
\newcommand{\cmark}{\ding{51}}%
\newcommand{\xmark}{\ding{55}}%

\title{Domain Generalization for Crop Segmentation with Standardized Ensemble Knowledge Distillation}
\date{}

\author{
  \href{https://orcid.org/0000-0002-4445-9783}{\includegraphics[scale=0.06]{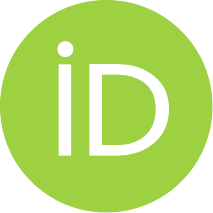}\hspace{1mm}Simone Angarano}\\ 
  Department of Electronics and Telecommunications \\ 
  Politecnico di Torino\\ 
  Turin, Italy \\ 
  \texttt{simone.angarano@polito.it} \\
  \And
  \href{https://orcid.org/0000-0002-6204-3845}{\includegraphics[scale=0.06]{orcid.pdf}\hspace{1mm}Mauro Martini} \\ 
  Department of Electronics and Telecommunications \\ 
  Politecnico di Torino\\ 
  Turin, Italy \\ 
  \texttt{mauro.martini@polito.it} \\
  \And
  \href{https://orcid.org/0009-0001-3768-3832}{\includegraphics[scale=0.06]{orcid.pdf}\hspace{1mm}Alessandro Navone} \\ 
  Department of Electronics and Telecommunications \\ 
  Politecnico di Torino \\ 
  Turin, Italy \\ 
  \texttt{alessandro.navone@polito.it} \\
  \And
  \href{https://orcid.org/0000-0002-1921-0126}{\includegraphics[scale=0.06]{orcid.pdf}\hspace{1mm}Marcello Chiaberge} \\
  Department of Electronics and Telecommunications \\ 
  Politecnico di Torino\\ 
  Turin, Italy \\ 
  \texttt{marcello.chiaberge@polito.it}
  }

\renewcommand{\headeright}{}
\renewcommand{\undertitle}{}
\renewcommand{\shorttitle}{Domain Generalization for Crop Segmentation with Standardized Ensemble Knowledge Distillation}

\hypersetup{
pdftitle={Domain Generalization for Crop Segmentation with Standardized Ensemble Knowledge Distillation},
pdfsubject={Domain Generalization for Crop Segmentation with Standardized Ensemble Knowledge Distillation},
pdfauthor={Simone Angarano, Mauro Martini, Alessandro Navone, Marcello Chiaberge},
pdfkeywords={Semantic Segmentation, Knowledge Distillation, Precision Agriculture},
}

\begin{document}
\maketitle

\abstract
In recent years, precision agriculture has gradually oriented farming closer to automation processes to support all the activities related to field management. Service robotics plays a predominant role in this evolution by deploying autonomous agents that can navigate fields while performing tasks such as monitoring, spraying, and harvesting without human intervention. To execute these precise actions, mobile robots need a real-time perception system that understands their surroundings and identifies their targets in the wild. Existing methods, however, often fall short in generalizing to new crops and environmental conditions. This limit is critical for practical applications where labeled samples are rarely available. 
In this paper, we investigate the problem of crop segmentation and propose a novel approach to enhance domain generalization using knowledge distillation. In the proposed framework, we transfer knowledge from a standardized ensemble of models individually trained on source domains to a student model that can adapt to unseen realistic scenarios.
To support the proposed method, we present a synthetic multi-domain dataset for crop segmentation containing plants of variegate species and covering different terrain styles, weather conditions, and light scenarios for more than 70,000 samples. We demonstrate significant improvements in performance over state-of-the-art methods and superior sim-to-real generalization. Our approach provides a promising solution for domain generalization in crop segmentation and has the potential to enhance a wide variety of agriculture applications.

%\keywords{Domain Generalization, Semantic Segmentation, Knowledge Distillation.}

\section{Introduction}
\label{introduction}
\begin{figure*}[t]
    \centering
    \includegraphics[width=\textwidth]{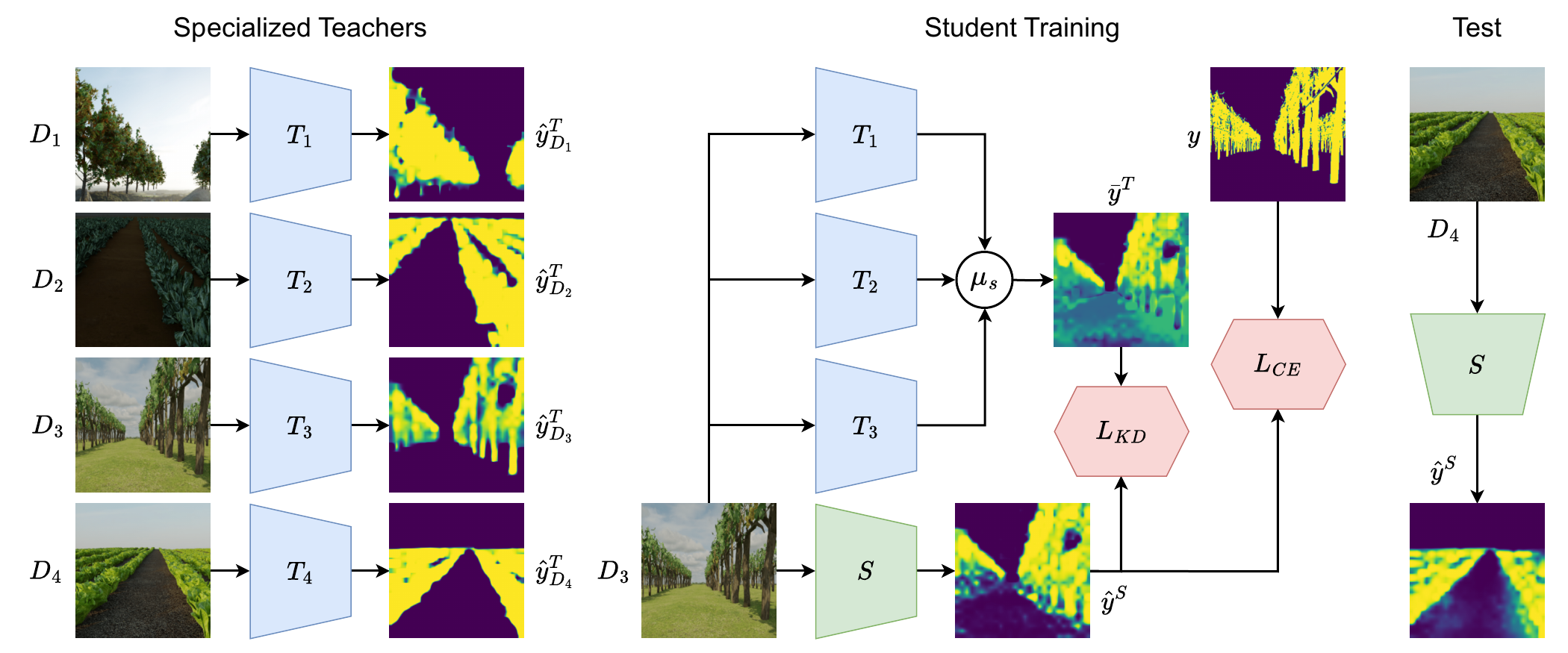}
    \caption{Schematic representation of the proposed distillation methodology for crop segmentation. Ensembled specialized teachers allow the student to obtain a standardized distillation mask ($\tilde y^T$) that is much more informative than the hard label ($y$) for robust student training. $\mu_s$ represents standardized ensembling.}
    \label{fig:scheme}
\end{figure*}
% CONTEXT
In the last two decades, scientific research in precision agriculture has significantly evolved its automatic and self-managed processes. Automation has been analyzed through four essential requirements: increasing productivity, allocating resources reasonably, adapting to climate change, and avoiding food waste \cite{zhai2020decision}. Recently, deep learning solutions led to new technological trends in all these tasks, providing competitive advantages for crop monitoring and managing \cite{ren2020survey}. Autonomous robots equipped with perception systems can assist or replace human operators in agricultural tasks such as harvesting \cite{harvesting}, spraying \cite{deshmukh2021design}, and vegetative assessment \cite{feng2020yield}, reducing human labor and enhancing operational safety. %These systems offer the potential to address these problems by perceiving the environment around them and acting accordingly. 
Various computer vision methods have been proposed for navigating and monitoring row crops, most of which are based on semantic segmentation \cite{luo2023semantic}.
Real-time crop segmentation can be used to identify objects on different scales: detailed leaf disease \cite{mukhopadhyay2021tea}, single fruits or branches \cite{peng2020semantic}, crop rows \cite{aghi2021deep}, and entire fields \cite{raei2022deep}.
It has also been exploited for autonomous navigation \cite{aghi2021deep}, combined with waypoint generation \cite{salvetti2023waypoint} or sensorimotor agents \cite{martini2022position}. 

% CHALLENGES
However, crop segmentation presents two main challenges. 
First, changing weather, lighting, terrain, and crop types pose a major obstacle to generalization. Supervised training methodologies usually reach remarkable results in well-defined experimental settings but struggle to yield good results where the data distribution changes \cite{csurka2017domain}. %Available deep learning models easily fail in realistic applications due to their ineffective generalization ability, blocking the development of truly autonomous systems \cite{dai2018dark,volk2019towards}. 
However, robustness in realistic scenarios can be enhanced using frameworks like domain generalization (DG). DG is a set of representation learning techniques that aims to train models capable of generalizing to unseen domains, i.e., out-of-distribution data.
Several DG methodologies have been presented in the last years, although often limiting their scope to classification on toy datasets \cite{zhou2022domain}. Applying generalization methods to realistic tasks is still limited to a few attempts \cite{choi2021robustnet,lee2022wildnet,martini2021domain}. Moreover, the considered domains are often limited to stylistic changes, overlooking more radical correlation shifts \cite{ye2022ood}. For instance, in some scenarios, brown and green positively correlate with terrain and vegetation. However, other domains present brown tree trunks and grass on the ground, inverting the correlation.

A second challenge is data availability, as no comprehensive dataset for crop segmentation across multiple scenarios is available. The reason is that on-field data collection and labeling are highly time-demanding. Hence, the only publicly available datasets focus on specific scenarios and usually include a modest number of samples. The scarcity of task-specific labeled data has recently favored the practice of synthetic data generation, leading to an additional Simulation-to-Reality (Sim2Real) gap and further compromising generalization \cite{barth2018data}.

% CONTRIBUTIONS
This work aims to effectively enhance DG in crop segmentation, working towards having a single model that can generalize across different crop types and environmental conditions.
It is well known that supervised neural networks exploit spurious correlations to find shortcuts in data and efficiently minimize the loss function \cite{geirhos2020shortcut}. In agricultural scenarios, these correlations can easily be found in the color of a specific species, low-level terrain textures, or background.
We apply the DG framework to encourage models to learn deep, robust features without knowing the target data distribution.
Moreover, recent findings have given a theoretical interpretation of the efficacy of model ensembling and knowledge distillation (KD) for robust representation learning \cite{allen2022towards}. Multiple features exist in data samples that can be used to classify them correctly, and this multi-view structure constitutes the "dark knowledge" that ensembles and KD exploit, explaining the efficacy of these methods. We investigate whether such property enhances domain and Sim2Real generalization, particularly for crop segmentation.

The proposed method distills knowledge from an ensemble of models individually trained on source domains to a student model that can adapt to unseen target domains, as depicted in Figure \ref{fig:scheme}. To effectively balance the contribution of the teachers, we standardize their output logits and, in this way, avoid overconfident predictions to guide knowledge transfer.
To properly validate the proposed method, we present the synthetic multi-domain dataset for crop segmentation \textsc{AgriSeg}, containing 11 crop types and covering different terrain styles, weather conditions, and light scenarios for more than $70,000$ samples. We conduct thorough experiments on \textsc{AgriSeg} and additional real-world datasets to verify the effectiveness of our method compared to other state-of-the-art solutions. %This study aims to extend the generalization of segmentation models to different background scenes and conditions. The models should identify the crops in the images without focusing on strong visual biases in the training data. This work and the proposed benchmark also pave the way for a unique crop segmentation model, achieving satisfying results among different crops.
The contributions of this work can be summarized as follows:
\begin{enumerate}
    \item We propose a novel DG methodology for crop segmentation based on ensemble KD weighted by logit standardization;
    \item We support our solution with \textsc{AgriSeg}, a rich multi-domain synthetic dataset to benchmark generalization in crop segmentation;
    \item We extensively experiment on synthetic and real data to demonstrate the improvement of the proposed method on state-of-the-art solutions.
\end{enumerate}
\noindent The code\footnote{\url{https://github.com/PIC4SeR/AgriSeg}} used for the experiments and the \textsc{AgriSeg} dataset\footnote{\url{https://pic4ser.polito.it/agriseg/}} are publicly available. % \url{https://github.com/PIC4SeR/AgriSeg} % \url{https://pic4ser.polito.it/agriseg/}

\section{Related Works}

\begin{figure*}[t]
\centering
\resizebox{\textwidth}{!}{
\begin{tabular}{cccc}
    \includegraphics[trim={0 0 0 1mm},clip,height=0.2\textwidth]{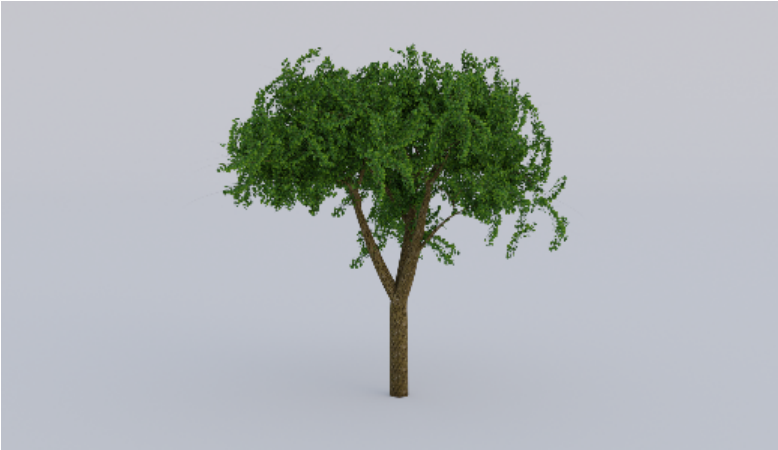} &
    \includegraphics[height=0.2\textwidth]{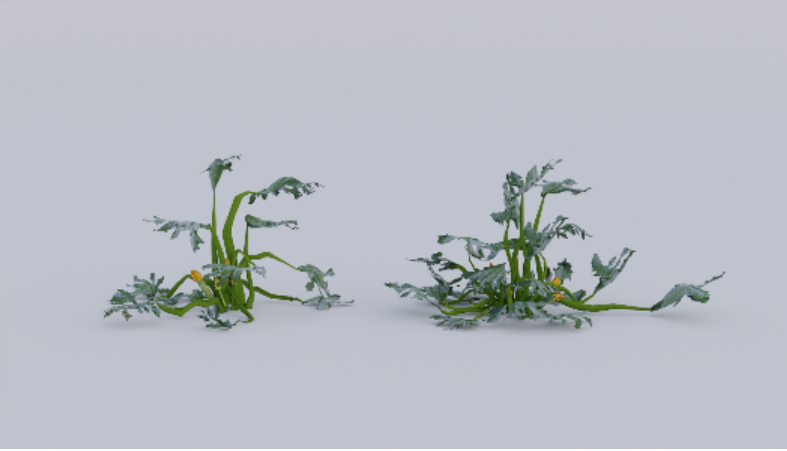} &
    \includegraphics[height=0.2\textwidth]{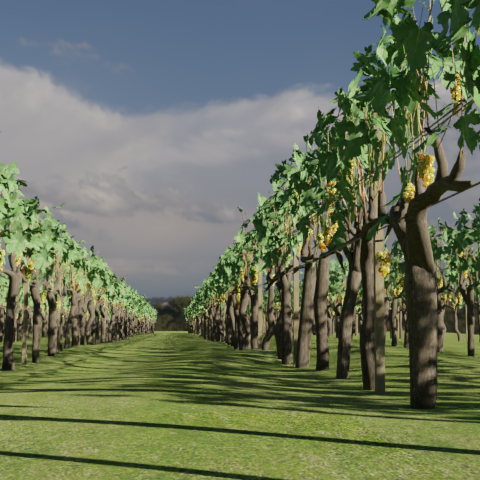} &
    \includegraphics[height=0.2\textwidth]{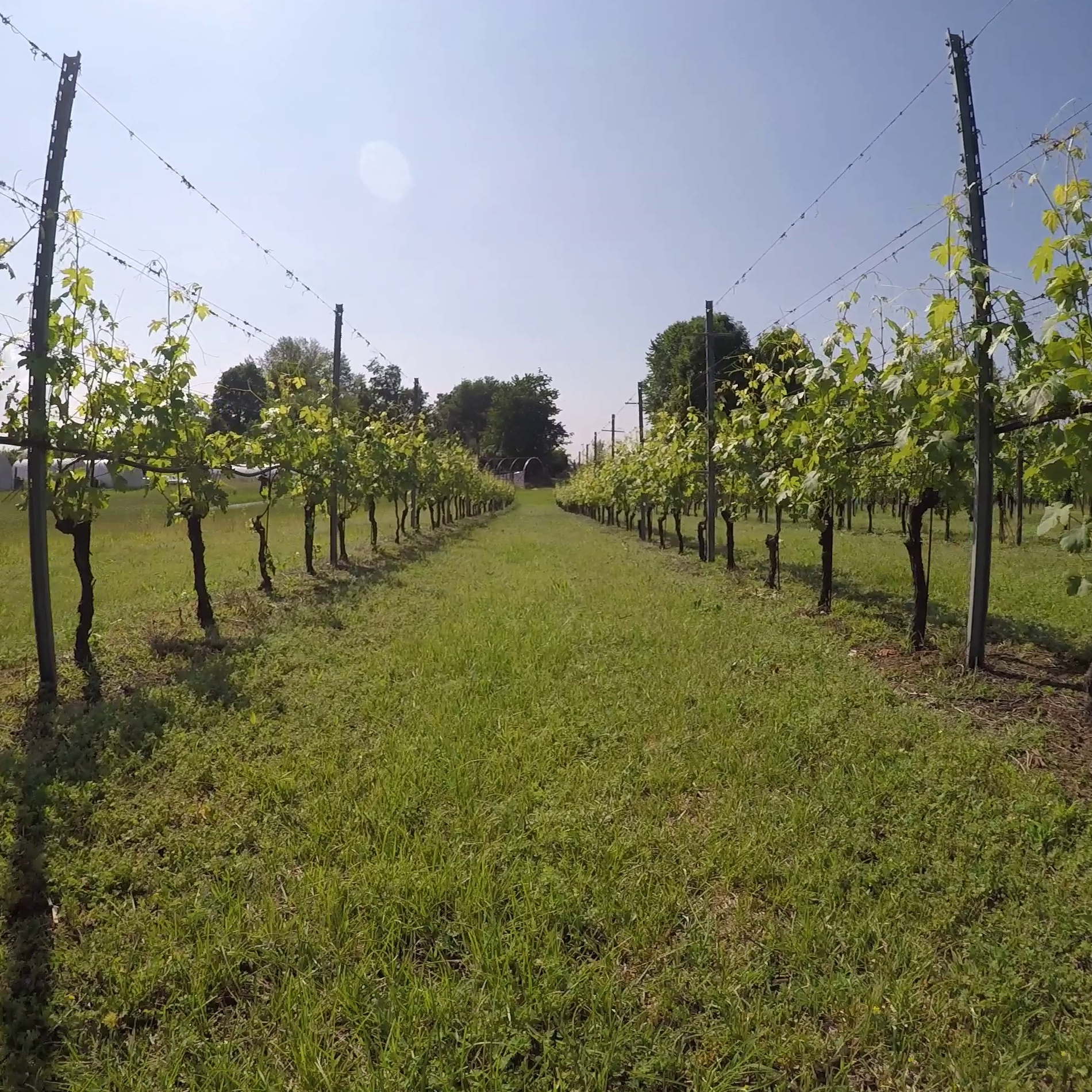}
    \\
    \includegraphics[height=0.2\textwidth]{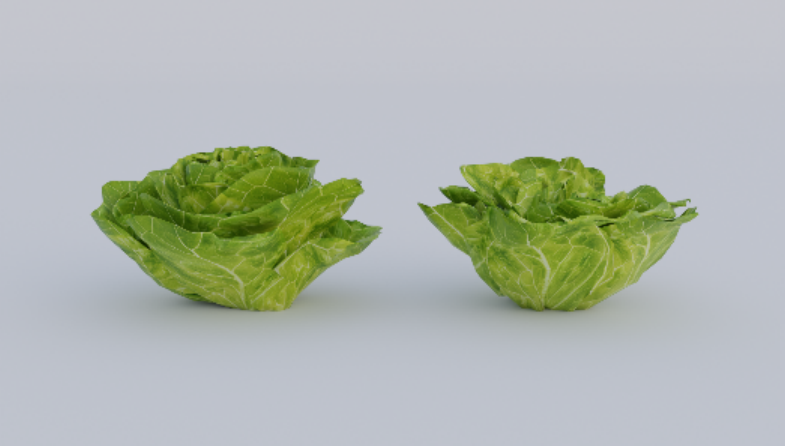} & 
    \includegraphics[trim={0 0 0 1mm},clip,height=0.2\textwidth]{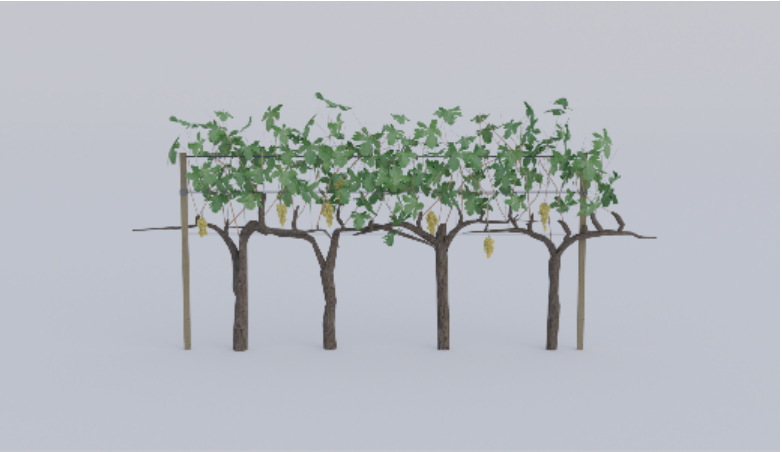} &
    \includegraphics[height=0.2\textwidth]{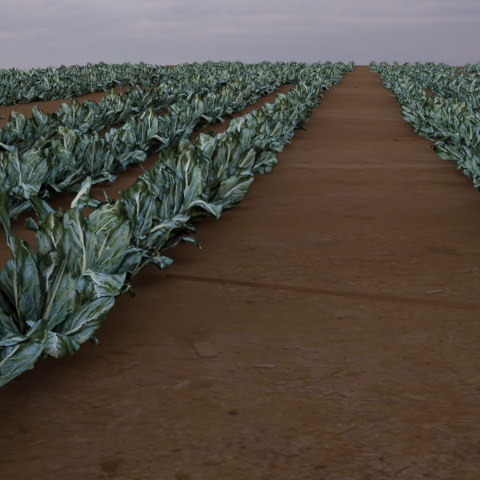} &
    \includegraphics[height=0.2\textwidth]{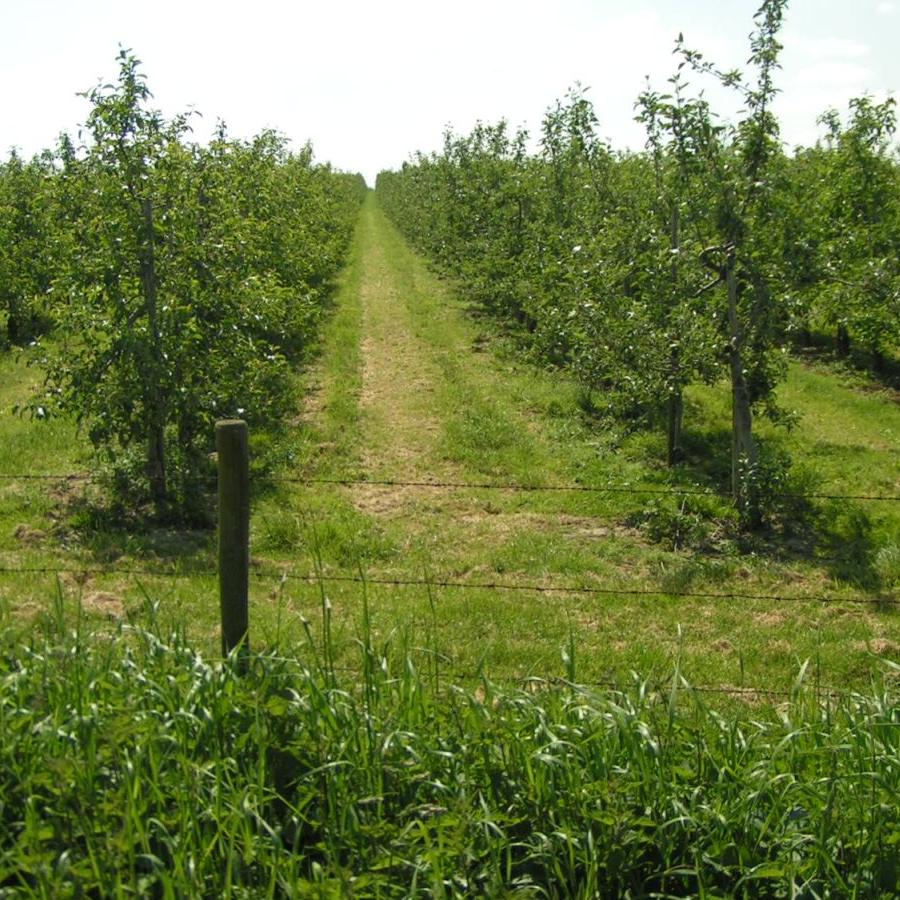}
\end{tabular}
}
\caption{From left to right: examples of synthetic 3D crop models used to build the \textsc{AgriSeg} Dataset (generic tree, zucchini, lettuce, vineyard); examples of resulting dataset images (vineyard, chard); examples of real-world test images (vineyard, miscellaneous).}
\label{fig:crop_models}
\end{figure*}

Generalization to Out-of-domain (OOD) data distributions is one of the most critical requirements for real-world computer vision applications like crop segmentation. Recently, rigorous validation benchmarks have been proposed to compare the advantages of different approaches and backbones for classification \cite{gulrajani2020search,angarano2022back}. In the meantime, segmentation across multiple scenarios has been studied, either designing massive foundation models \cite{kirillov2023segany} or creating new DG methods. As we aim to push the limits of generalization for efficient and easily deployable architectures, we focus on the latter approach. In particular, \cite{pan2018two} proposed an Instance Batch Normalization (IBN) block for residual modules to avoid bias toward low-level domain-specific features like color, contrast, and texture. \cite{nuriel2021permuted}, on the same line, proposed a permuted Adaptive Instance Normalization (PAdaIN) block, which works at both low-level and high-level features, randomly swapping second-order statistics between source domains and hence regularizing the network towards invariant features. \cite{choi2021robustnet} proposed RobustNet, a model incorporating an Instance Selective Whitening (ISW) loss disentangling and removing the domain-specific style in feature covariance. \cite{lee2022wildnet} proposed to extract domain-generalized features by leveraging a variety of contents and styles using a "wild" dataset.
Most recently, \cite{wang2021embracing} has been the first attempt to apply KD in the DG framework for classification tasks, proposing a gradient filtering approach. \cite{lee2022cross} proposed Cross-domain Ensemble Distillation (XDED) to extract the knowledge from domain-specific teachers and obtain a general student. However, this setup was only applied to classification, while the authors used a different approach for segmentation distilling from a single teacher.
Standard DG benchmarks almost solely focus on domestic environments or autonomous driving \cite{pascal-voc-2012,cordts2016cityscapes}, and generalization for crop segmentation has been addressed only in the last few years. In particular, \cite{weyler2023towards} proposes a style transfer method for robust weed segmentation, considering only one crop type. \cite{Panda_2023_CVPR} proposes supervised Domain Adaptation for row crop segmentation, requiring target-domain labeled data. We push the generalization concept further, including not only weather and lighting conditions but also aiming to generalize to unseen crop types without prior knowledge about the target data distribution. We take advantage of the capabilities provided by ensemble KD \cite{allen2022towards} to transfer the knowledge of domain-expert teachers to a general multi-domain student. We improve the method proposed in \cite{lee2022cross} for classification, adding logit standardization to balance the contribution of different teachers to the KD loss and applying it to real-world crop segmentation. Each teacher must be trained only once, and the method can be extended to more domains by just training a new teacher and then distilling.

\section{Methodology}
\label{sec:methodology}
\begin{comment}
In this section, we describe the proposed methodology for DG in the crop segmentation task. 
First, we theoretically define the problem of DG and its application to semantic segmentation. 
% Secondly, we briefly review the adopted architecture (LR-ASPP).
Then, we describe the proposed training procedure in detail, which combines standard empirical risk minimization with an auxiliary loss derived by ensemble KD. 
Finally, we define the logit standardization mechanism we added to balance the contribution of different teachers and discourage the model from being biased by low-level domain-specific features.
\end{comment}

\subsection{Domain Generalization}
\label{sec:dg}
Given the input random variable \(X\) with values \(x \in \mathcal{X}\) and the target random variable \(Y\) with values \(y \in \mathcal{Y}\), the definition of \textit{the domain} is associated with the joint probability distribution \(P(X,Y)\) (\(P_{XY}\) for simplicity) over \(\mathcal{X}\)$\times$\(\mathcal{Y}\). Supervised learning aims to train a classifier \(f:\mathcal{X} \to \mathcal{Y}\) exploiting $N$ available labeled examples of a dataset \(D = {(x_i, y_i)}^N_{i=1}\) that are identically and independently distributed (i.i.d.) and sampled according to \(P_{XY}\). The goal of the training process is to minimize the \textit{empirical risk} associated with a loss function \(l:\mathcal{Y}\times\mathcal{Y} \to [0,+\infty)\), 
\begin{equation}
    R_{\text{emp}}(f)={\frac {1}{N}}\sum _{i=1}^{N}l(f(x_{i}),y_{i})
\end{equation}

\noindent by learning the classifier \(f\). Dataset $D$ is the only available source of knowledge to learn \(P_{XY}\). We refer to this basic learning method as empirical risk minimization (ERM) \cite{vapnik1999overview} and use it as a baseline for the experimentation. 

In DG, a set of different \(K\) source domains \(\mathcal{S}=(S_k)^K_{k=1}\) is used to learn a classifier \(f\) that aims at generalizing well on an unknown target domain \(T\). Each source domain is associated with its joint probability distribution \(P_{XY}^k\), whereas \(P_{XY}^\mathcal{S}\) indicates the overall source distribution learned by the classifier \cite{zhou2022domain}. Indeed, DG aims to enable the classifier to predict well on out-of-distribution data, namely on the target domain distribution \(P_{XY}^T\), by learning an overall domain-invariant distribution from the source domains seen during training.

\subsection{Knowledge Distillation}
\label{sec:kd}
KD aims at transferring the knowledge learned by a \textit{teacher} model to a smaller or less expert \textit{student} model. It was first proposed in \cite{buciluǎ2006model}, received greater attention after \cite{hinton2015distilling}, and is one of the most promising techniques for model compression and regularization today. In its original formulation based on classification, KD applies an auxiliary loss to the output logits of the student $z_S(x)\in \mathbb R^C$, where C is the number of classes. The posterior predictive distribution of $x$ can be formulated as:
\begin{equation}
    P(y|x;\theta,\tau) = \frac{exp(z_y(x)/\tau)}{\sum_{i=1}^{C}exp(z_i(x)/\tau)}
\end{equation}
\noindent where $y$ is the label, $\theta$ is the set of parameters of the model, and $\tau$ is the temperature scaling parameter. To match the distributions of student and teacher, KD minimizes the Kullback-Leibler Divergence between the two:
\begin{align}
   &L_\text{KD}(X;\theta,\tau) = \sum_{x_i\in X} \sum^C_{c=1} L_\text{KD}^{x_i,c}\\ 
   &L_\text{KD}^{x_i,c} = D_\text{KL}(P(c|x_i;\theta_T,\tau)||P(c|x_i;\theta_S,\tau))
\end{align}
\noindent where $X$ is a batch of input samples and $\theta_T$ and $\theta_S$ are the parameters of teacher and student, respectively. 
In this work, we apply a novel KD technique for semantic segmentation to improve models' generalization ability across domains.

\subsection{Standardized Ensemble Distillation}
\label{subsec:waypoints}
We propose a simple yet effective training procedure based on model ensemble, KD, and logit standardization to encourage the model to learn domain-invariant features. We choose ensemble KD encouraged by the recent theory of \cite{allen2022towards} on multi-view extraction from data. Ensemble KD has been previously applied to classification in XDED \cite{lee2022cross}, leveraging the separate pretraining of a teacher for each source domain and distilling the ensembled predicted logits. We aim to apply the same intuition to crop semantic segmentation, taking into account the differences between the two tasks and the additional challenges given by the agricultural setting and the Sim2Real gap. Another important challenge of this application scenario is that domain shifts are not only given by style transfer but also by the presence of completely different crop types. %As a remark, the authors of XDED also proposed a semantic segmentation method in the same paper but without involving model ensembling. In particular, they instead average all the output logits in a training batch that correspond to the same ground-truth label. We compare with XDED in \ref{sec:results}, observing a considerable performance gap.

\begin{figure}[t]
    \centering
    \resizebox{0.9\columnwidth}{!}{%    
    \begin{tabular}{ccccc}
    & \textbf{Gen. Tree 2} & \textbf{Chard} & \textbf{Lettuce} & \textbf{Vineyard} \\
     \rotatebox[origin=c]{90}{\textbf{ERM}} & \includegraphics[height=.3\linewidth,valign=m]{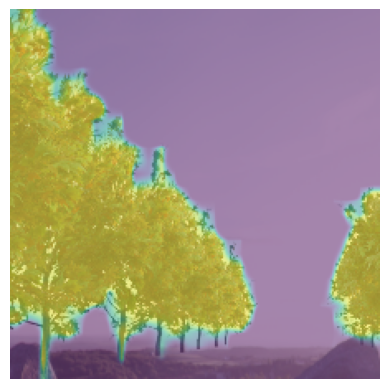} & 
    \includegraphics[height=.3\linewidth,valign=m]{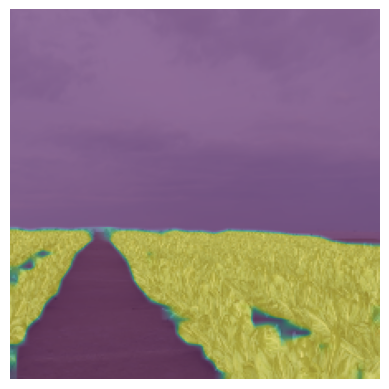} & 
    \includegraphics[height=.3\linewidth,valign=m]{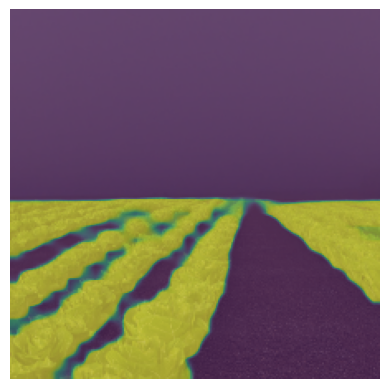} & 
    \includegraphics[height=.3\linewidth,valign=m]{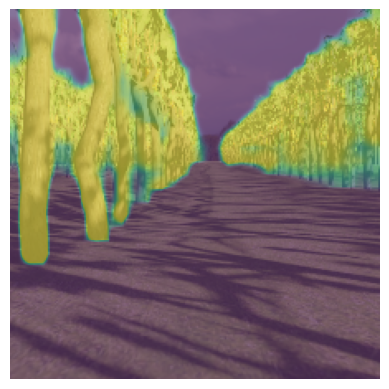} \\
    \rotatebox[origin=c]{90}{\textbf{Ensemble}} & \includegraphics[height=.3\linewidth,valign=m]{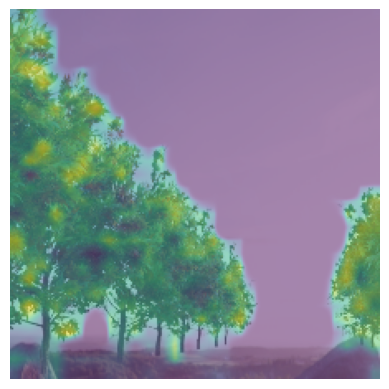} & 
    \includegraphics[height=.3\linewidth,valign=m]{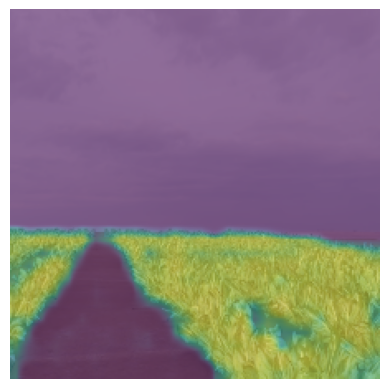} & 
    \includegraphics[height=.3\linewidth,valign=m]{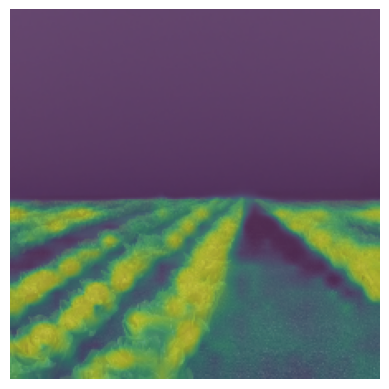} & 
    \includegraphics[height=.3\linewidth,valign=m]{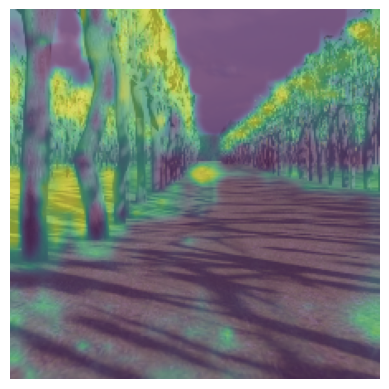}
    \end{tabular}
    }
    \caption{Comparison of ERM predictions with our ensemble of specialized teachers. While for simpler domains, the predictions of the specialized teachers agree and return a high-confidence mask, for challenging ones, the teachers give an uncertain but more informative mask that can be distilled into the student.}
    \label{fig:ensemble}
\end{figure}

In our proposed method, we train a teacher for each source domain and ensemble them to create the distillation knowledge:
\begin{equation}
    \bar y^T(x) = \frac{1}{D}\sum^D_{d=1} \hat{y}^T_d(x)
\end{equation}
\noindent where $\hat{y}^T_d(x)$ is the predicted logits tensor for the source domain $d$, $\bar y^T$ is the ensembled teacher logits tensor, and $D$ is the number of source domains.
The motivation behind this choice is that by averaging the predictions of different specialized models, the resulting map is much more informative than the ground-truth label. As depicted in figure \ref{fig:ensemble}, the teacher's segmentation is less confident and often assigns non-zero probabilities to disturbing elements such as grass and background vegetation. This spurious information guides the student towards implicitly recognizing what features are more likely to be confounding at test time. This information does not overcome label supervision, as the distillation loss has a relatively low weight in the optimization process. On the contrary, if the distillation mask is very confident, the student is guided toward being more confident and implicitly incorporates the information that a certain domain is easier to segment. This effect can be tuned using a temperature factor and a weight loss.

We train the student in the standard ERM DG framework with an additional distillation loss based on the distance between the student logits and the ensembled teacher logits.
To improve the effectiveness of the distillation mechanism, we design a simple mechanism to prevent the student from being biased by the overconfidence of the teachers. Indeed, each teacher is trained on a single domain and hence can fall for spurious correlations in the training data (e.g., color bias). We hence propose to standardize teacher and student logits \cite{sun2024logit} before distilling as follows:
\begin{equation}
    \tilde y^T = \frac{y^T-\bar y^T}{\sigma(y^T)\cdot\tau}, \qquad \tilde y^S = \frac{\hat y^S-\bar y^S}{\sigma(\hat y^S)\cdot\tau}
\end{equation}
\noindent where $\bar y$ is the mean, $\sigma(y)$ is the standard deviation of the logits, and $\tau$ is the temperature. The intuition behind this choice is that optimal $\tau$ could vary across domains due to the teachers being more or less confident about their predictions. Standardization allows for adaptive calibration of logit temperature and effective domain knowledge transfer.
% We leverage the recent findings by \cite{Shu_2021_ICCV} and modify the distillation loss function to exploit the channel-wise information extracted from the network. 

% In particular, we apply the softmax operator $\phi$ along the flattened spatial dimension instead of the channel dimension before computing the loss:
% \begin{equation}
%     \phi(\hat y^S) = \frac{exp(\hat y_i^S/\tau)}{\sum_{i=1}^{W\cdot H}exp(\hat y_i^S/\tau)}
% \end{equation}
% \noindent where $\hat y_i^S$ is the $i$-th element of the flattened student logit tensor $\hat y^S$, $W\cdot H$ is its spatial dimension, and $\tau$ is the temperature. The same operation is applied to the teacher logits $\bar y^T$.
%\begin{comment}
The distillation loss is calculated as the Kullback-Leibler divergence between teacher and student logits:
\begin{equation}
    L_\text{KD}(\tilde y^T,\tilde y^S) = \frac{\tau^2}{C} \sum^C_{c=1}\sum^{W\cdot H}_{i=1} \phi(\tilde y^T_{c,i})\cdot log\biggl(\frac{\phi(\tilde y^T_{c,i})}{\phi(\tilde y^S_{c,i})}\biggr) 
\end{equation}
\noindent where $C$ is the number of classes and $\tau$ is the temperature. %For the specific case of binary segmentation, the formulation is simplified as the predicted mask consists of only one channel ($C=1$).

In combination with the distillation loss, we optimize the standard cross-entropy loss between student logits and ground-truth labels $y$:
\begin{equation}
    L_\text{CE}(y,\hat y^S) = -\sum_{i=1}^{C} y_i\cdot log(\hat y^S_i)
\end{equation}
\noindent which for binary segmentation becomes a simple binary cross-entropy loss. 
%\end{comment}

The overall loss can be written as follows:
\begin{equation}
    L(y,\bar y^T,\hat y^S) = L_\text{CE}(\bar y,\hat y^S) + \lambda L_\text{KD}(\tilde y^T,\tilde y^S)
\end{equation}
\noindent where $\lambda$ is a weighting parameter to balance the loss components. We remark that our method adds no overhead at test time. We provide a thorough ablation of the various components of our method in \ref{subsec:ablation} to highlight the strong improvement on previous solutions.

% \begin{figure*}[ht]
%     \centering
%     \includegraphics[width=\textwidth]{media/architecture.pdf}
%     \caption{Architecture of the adopted network with MobileNetV3 as the backbone and LR-ASPP as the head\cite{howard2019searching}. Below each block, we report the spatial scaling factor of the features compared to the input size. The batch dimension is omitted for clearness.}
%     \label{fig:architecture}
% \end{figure*}

\setlength{\tabcolsep}{5pt}
\begin{table}[t]
    \centering
    \begin{tabular}{l c c c c}
    \toprule
    \textbf{Crop}               & \textbf{Samples}       & \textbf{Type}      & \textbf{Category↓} & \textbf{Height {[}m{]}} \\
    \midrule
    \textit{Lettuce}             & 4800          & Synthetic & Low      & 0.22                   \\
    \textit{Chard}               & 4800          & Synthetic & Low      & 0.25                   \\
    \textit{Lavender}            & 5260          & Synthetic & Low      & 0.3                    \\
    \textit{Zucchini}            & 19200         & Synthetic & Medium   & 0.6                    \\
    \textit{Cotton}              & 4800          & Synthetic & Medium   & 0.6                    \\
    \textit{Vineyard}            & 4800          & Synthetic & Tall     & 1.5-2.5                    \\
    \textit{Pergola Vineyard}    & 4800          & Synthetic & Tall     & 3.2                    \\
    \textit{Apple Tree}          & 9600          & Synthetic & Tall     & 2.7                    \\
    \textit{Pear Tree}           & 4800          & Synthetic & Tall     & 3.0                   \\
    \textit{Generic Tree 1}      & 4800          & Synthetic & Tall     & 4.5                    \\
    \textit{Generic Tree 2}      & 2785          & Synthetic & Tall     & 4.5                    \\
    \midrule
    \textit{Vineyard} \cite{aghi2021deep} & 500  & Real      & Tall     & 2.5                    \\ 
    \textit{Miscellaneous}       & 100           & Real      & Any      & Any                    \\ 
    \textit{VegAnn} \cite{madec2023}              & 3775          & Real      & Any        & Any                      \\ 
    \bottomrule
    \end{tabular}
    \vspace{5pt}
    \caption{Detailed properties for each domain of the \textsc{AgriSeg} dataset. The section on the top reports the synthetic crop datasets generated in simulation, while the section on the bottom the real-world ones.}
    \label{tab:dataset}
    
\end{table}

\section{Experimental Setting}
\label{sec:experimental}
This section describes the details of the proposed synthetic \textsc{AgriSeg} segmentation dataset and the procedure we followed to validate the effectiveness of our DG methodology. 
In \ref{subsec:dataset}, we review the procedure followed to generate the \textsc{AgriSeg} dataset, while in section \ref{subsec:training}, details on the training framework and implementation are given.

\subsection{Dataset}
\label{subsec:dataset}
To generate the synthetic crop dataset with realistic plant textures and measurements, high-quality 3D plant models have been created using Blender\footnote{\url{https://www.blender.org/}}. A wide variety of crops have been included in the dataset to validate the segmentation performance of the model trained with the proposed DG method. Depending on the plant's height, three primary macro-categories of crops have been identified. Low crops, such as lettuce and chard, have an average height of 20-25 cm. Medium crops, such as zucchini, grow to 60 cm. Tall crops, which include vineyards and trees, can grow up to 2.5-4.5 m. %A meaningful target performance to be achieved by the segmentation model is set to generalize to previously unseen plant species inside the same macro-category, which differ mostly in the color features and slightly in the geometrical shape. 
Some examples of 3D plant models are shown in Figure \ref{fig:crop_models}.

\begin{table*}[t]
\centering
\begin{tabular}{ccccc|c}
\toprule 
\textbf{Method} & \textbf{Gen. Tree 2} & \textbf{Chard} & \textbf{Lettuce}  & \textbf{Vineyard} & \textbf{Average} \\ \midrule
Teacher & $84.52 \pm 0.62$ & $95.09 \pm 0.11$ & $95.37 \pm 0.10$ & $85.51 \pm 1.08$ & $90.12 \pm 0.48$ \\ \midrule
ERM\cite{vapnik1999overview} & $76.31 \pm 1.53$   & $87.63 \pm 0.86$ & $80.64 \pm 5.00$ & $67.34 \pm 2.18$ & $77.98 \pm 1.62$ \\
IBN\cite{pan2018two} & $79.15 \pm 1.57$   & $88.92 \pm 1.07$ & $57.85 \pm 6.00$ & $69.11 \pm 2.88$ & $73.76 \pm 2.36$ \\
ISW\cite{choi2021robustnet} & $77.14 \pm 1.77$   & $89.44 \pm 0.47$ & $53.86 \pm 8.45$ & $68.76 \pm 3.06$ & $72.30 \pm 2.30$ \\
pAdaIN\cite{nuriel2021permuted} & $75.17 \pm 2.03$   & $86.65	\pm 1.75$ & $78.05 \pm 4.90$ & $69.79\pm 1.41$ & $77.41 \pm 0.77$ \\
WildNet\cite{lee2022wildnet} & $82.34 \pm 1.55$   & $93.68 \pm 0.08$ & $43.55 \pm 6.91$ & $72.83 \pm 0.76$ & $73.10 \pm 2.70$ \\ 
CWD\cite{shu2021channel} & $ 64.20 \pm 8.08 $ & $ 84.70 \pm 1.92 $ & $ 83.84 \pm 2.79 $ & $ 62.88 \pm 2.49 $ & $ 73.90 \pm 2.17 $ \\
WCTA\cite{weyler2023towards} & $75.09 \pm 0.94$ & $86.66 \pm 1.93$ & $70.73 \pm 10.85$ & $66.57 \pm 3.20$ & $74.76 \pm 2.25$ \\ 
KDDG\cite{wang2021embracing} & $80.13 \pm 1.61$ & $87.67 \pm 1.66$ & $74.16 \pm 3.37$ & $65.55 \pm 1.18$ & $76.88 \pm 0.68$\\ \midrule
XDED\cite{lee2022cross} & $77.18 \pm 2.20$   & $88.62 \pm 0.71$ & $75.82 \pm 5.12$ & $70.48 \pm 1.00$ & \underline{$78.03 \pm 1.85$}\\
\textbf{Ours} & $78.84 \pm 1.24$   & $88.35 \pm 1.27$ & $78.04 \pm 3.46$ & $72.21 \pm 1.02$ & $\pmb{79.36 \pm 1.04}$ \\ \bottomrule
\end{tabular}
\vspace{5pt}
\caption{Comparison between the proposed methodology and other state-of-the-art DG algorithms for semantic segmentation adopting the leave-one-out DG validation procedure described in \ref{subsec:training}. We report the Intersection-over-Union (IoU) metric (in \%) for each result as mean and standard deviation. The best and second-best overall results are highlighted and underlined, respectively.}
\label{tab:dg}
\end{table*}

Various terrains and sky models have been used to achieve realistic background and light conditions. %The generalization properties of the segmentation network are enhanced considering the light of different moments of the day and various weather conditions. 
Afterward, Blender's Python scripting functionality was used to automatically separate plants from the rest of the frame and generate a dataset of RGB images and corresponding segmentation masks. This work presents the \textsc{AgriSeg} dataset, composed of samples from low crops (e.g., chard and lettuce), medium crops (e.g., zucchini), and tall crops (e.g., vineyard, pear tree, and generic tall tree). Each dataset presents four sub-datasets that differ in background and terrain. Cloudy and sunny skies, diverse lighting, and shadow conditions are included. Camera position and orientation have been changed to capture diverse image samples along the whole field for each sub-dataset. Overall, the \textsc{AgriSeg} dataset contains more than $70,000$ samples. In the bottom rows, we also include three additional domains to validate the considered solutions on real-world data as a final test. The \textit{Real Vineyard} dataset was originally presented in \cite{aghi2021deep}, but the proposed labels were coarse. Hence, we re-label the samples using the \textit{SALT} labeling tool \footnote{\url{https://github.com/anuragxel/salt}} based on Segment Anything \cite{kirillov2023segany}. We also add \textit{Miscellaneous}, containing 100 samples from disparate crop types, and label it similarly. Finally, we include \textit{VegAnn} \cite{madec2023}, a multi-crop dataset acquired under diverse conditions for vegetation segmentation. This domain constitutes a highly different setting from the training domains, so we use it to evaluate generalization in extreme domain shifts. Details for each dataset are listed in Table \ref{tab:dataset}.

\subsection{Training}
\label{subsec:training}
In this section, we report all the relevant information regarding experimental settings for model training and testing: data preprocessing, hyper-parameter search, and implementation.
We repeat each training five times with different and randomly generated seeds to obtain statistically relevant metrics. %In this way, neither hyper-parameter search nor benchmarks can take advantage of trial repeatability, as data splitting, augmentation, and weight initialization change from one iteration to the next. 
Each of our benchmark's results is reported as mean and standard deviation. 

\subsubsection{Data Preprocessing}
\label{subsec:data_aug}
We preprocess input images through the ImageNet standard normalization \cite{imagenet} to use pretrained weights. We apply the same data augmentation to all the experiments, consisting of random cropping with a factor in the range $[0.5,1]$ and flipping with a probability of 50\%. We don't use random jitter, contrast, and grayscale, which are common practices in DG. We instead draw inspiration for the WCTA stylization method proposed in \cite{weyler2023towards} for weed segmentation. We apply it randomly with probability $p=0.001$ and call our version pWCTA.
The reason is that we don't tackle just a shift in style but also in context (the change of crop type), and stronger stylization could over-regularize training. Experiments confirm that our choice leads to enhanced generalization on the proposed dataset.

\subsubsection{Hyper-parameters} 
\label{subsec:hyperparameters}
We conduct a random search to determine the optimal training hyper-parameters for the ERM DG baseline. We define a range of values for continuous arguments and a set of choices for discrete ones and select the best combination via the \textit{training-domain validation set} strategy proposed in \cite{gulrajani2020search}. It consists of picking the model that maximizes the metric (in our case, IoU) on a validation split of the training set (in our case, 10\%, uniform across domains) at the end of each epoch. %This selection method assumes that the average distribution of source domains is similar to that of the target domain on which the best model is tested. 

We choose a batch size $B=64$ and set the number of training epochs to 50. We choose temperature $\tau=2$ and weight $\lambda=0.1$ for $L_{KD}$. %Following the procedure proposed in \cite{lee2022cross}, we combine KD with feature whitening and apply UniStyle to the first layers of the backbone (results are reported in \ref{sec:results}). 
We use AdamW \cite{loshchilov2017decoupled} as the optimizer with a weight decay of $10^{-5}$. The learning rate is scheduled with a polynomial decay between $5\times10^{-5}$ and $5\times10^{-6}$. We compare to state-of-the-art methodologies running the same hyper-parameter search when tuning is necessary. We apply IBN \cite{pan2018two} and ISW\cite{choi2021robustnet} to the first two blocks of the backbone, while pAdaIN \cite{nuriel2021permuted} is applied to all the layers with a probability of $10^{-3}$. The ISW loss is weighted by a factor of $10^{-2}$, while XDED \cite{lee2022cross} is applied with a weight of $10^{-3}$, a $\tau$ of 2, and in combination with UniStyle feature whitening.

\begin{table*}[t]
\centering
\begin{tabular}{cccccc|c}
\toprule
\textbf{Method} & \textbf{Pear Tree} & \textbf{Zucchini} & \textbf{Real Vineyard} & \textbf{Real Misc.} & \textbf{VegAnn} & \textbf{Average} \\ \midrule 
Teacher & $90.84 \pm 0.30$ & $90.42 \pm 0.11$ & $69.20 \pm 2.86$ & $54.39 \pm 3.71$ & $ 85.70
\pm 0.88$ & $78.11 \pm 1.57$ \\ \midrule
ERM\cite{vapnik1999overview} & $82.11 \pm 0.93$ & $86.11 \pm 0.15$ & $51.51 \pm 7.27$ & $67.48 \pm 0.97$ & $61.92 \pm 0.93$ & $69.83 \pm 1.33$\\
IBN\cite{pan2018two} & $82.24 \pm 0.64$ & $86.06 \pm 0.07$ & $52.04 \pm 3.98$ & $67.97 \pm 2.21$ & $63.06 \pm 1.51$ & $70.27 \pm 1.68$ \\
ISW\cite{choi2021robustnet} & $82.31 \pm 0.83$ & $86.03 \pm 0.11$ & $55.46 \pm 2.83$ & $67.68 \pm 2.04$ & $63.29 \pm 1.42$ & $70.95 \pm 1.45$ \\
pAdaIN\cite{nuriel2021permuted} & $82.67 \pm 0.77$ & $85.96 \pm 0.20$ & $53.02 \pm 6.26$ & $63.39 \pm 2.00$ & $62.56 \pm 1.26$ & $69.52 \pm 2.10$ \\
WildNet\cite{lee2022wildnet} & $88.50 \pm 0.31$ & $86.32 \pm 0.10$ & $37.30 \pm 0.61$ & $72.15 \pm 0.27$ & $42.62 \pm 1.36$ & $65.38 \pm 0.53$ \\
CWD\cite{shu2021channel} & $ 79.52\pm 0.54$ & $85.84 \pm 0.07$ & $50.83 \pm 4.37$ & $65.27 \pm 3.28$ & $61.18 \pm 1.21$ & $68.53 \pm 1.89$ \\
WCTA\cite{weyler2023towards} & $81.80 \pm 0.82$ & $85.87 \pm 0.32$ & $54.83 \pm 1.50$ & $64.81 \pm 3.99$ & $63.22 \pm 2.17$ & $70.11 \pm 1.76$ \\
KDDG\cite{wang2021embracing} & $81.69 \pm 0.50$ & $86.22 \pm 0.13$ & $55.99 \pm 2.61$ & $62.60 \pm 4.23$ & $63.49 \pm 0.63$ & $70.00 \pm 1.62$ \\ \midrule
XDED\cite{lee2022cross} & $82.04 \pm 0.56$ & $86.11 \pm 0.25$ & $56.10 \pm 7.30$ & $66.92 \pm 1.93$ & $64.09 \pm 0.80$ & \underline{$71.05 \pm 1.41$} \\
\textbf{Ours} & $83.83 \pm 0.11$ & $86.39 \pm 0.04$ & $57.21 \pm 3.49$ & $69.84 \pm 1.34$ & $65.00 \pm 1.55$ & $\pmb{72.45 \pm 1.31}$ \\ 
\bottomrule
\end{tabular}
\vspace{5pt}
\caption{Comparison between the proposed methodology and other state-of-the-art DG algorithms on additional target domains. We train the models on all four domains chosen for the previous benchmark. We report IoU (in \%) on the unseen domains as mean and standard deviation. The best and second-best results are highlighted and underlined, respectively.}
\label{tab:test}
\end{table*}

\subsubsection{Implementation}
\label{subsec:implementation}
To tackle a realistic real-time application and following previous work on crop segmentation \cite{aghi2021deep}, we choose MobileNetV3 \cite{howard2019searching} with an LR-ASPP segmentation head\cite{howard2019searching} as the model architecture. This choice provides an optimal trade-off between performance and efficiency, exploiting effective modules such as depth-wise convolutions, channel-wise attention, and residual skip connections.
%Our experimentation code is developed in Python 3 using TensorFlow as the deep learning framework. 
We train models starting from ImageNet pretrained weights, so the input size is fixed to $(224, 224)$. The considered state-of-the-art DG methodologies are taken from the official repositories when available or reimplemented. All the training runs are performed on a single Nvidia RTX 3090 GPU.

\section{Results}
\label{sec:results}

In this section, we present the main results of the experimentation conducted to evaluate the effectiveness of the proposed methodology. First, we compare our distillation-based approach with recent and promising DG and semantic segmentation alternatives. Inspired by popular datasets for image classification, we select four domains (\textit{Generic Tree 2}, \textit{Chard}, \textit{Lettuce}, and \textit{Vineyard}) and evaluate all the methodologies by training on three domains and testing on the fourth. The domains are selected to cover different crop dimensions and visual characteristics and guarantee a challenging generalization benchmark. Then, we perform an additional evaluation by training models on all four datasets and testing on five additional target domains (\textit{Pear Tree}, \textit{Zucchini}, \textit{Real Vineyard}, \textit{Real Miscellaneous}, and \textit{VegAnn}). %We also report the predicted masks for a qualitative comparison on some random samples. 
In addition, we conduct a small ablation study to investigate the role of different components in our methodology and the importance of specialized single-domain teachers.

\begin{table*}[t]
\centering
\begin{tabular}{@{}cccc|ccc|c@{}}
\toprule
\textbf{Method} &
\textbf{Teacher} &
%\textbf{Style Aug} &
\textbf{pWCTA} &
\textbf{Logit Std} &
\textbf{Vineyard Real} &
\textbf{Misc. Real} &
\textbf{VegAnn} &
\textbf{Average} \\ \midrule
ERM & \xmark & \xmark &  \xmark & $51.51 \pm 7.27$ & $67.48 \pm 0.97$ & $61.92 \pm 0.93$ & $60.30 \pm 3.06$ \\
KD & ERM & \xmark & \xmark & $61.27 \pm 1.03$ & $63.70 \pm 1.72$ & $64.99 \pm 0.47$ & $63.32 \pm 1.07$ \\
XDED \cite{lee2022cross} & Ens. & \xmark & \xmark & $56.10 \pm 7.30$ & $66.92 \pm 1.93$ & $64.09 \pm 0.80$ & $62.37 \pm 3.34$ \\
\midrule
\multirow{2}{*}{\textbf{Ours}} 
& Ens. & 0.001 & \xmark & $57.01 \pm 2.53$ & $69.07 \pm 0.69$ & $65.36 \pm 1.12$  & \underline{$63.81 \pm 1.45$} \\
%& Ens. & 0.001 & norm & $49.50 \pm 9.88$ & $66.00 \pm 1.36$ & $63.35 \pm 0.72$ & $59.62 \pm 3.99$ \\
& Ens. & 0.001 & \cmark & $57.21 \pm 3.49$ & $69.84 \pm 1.34$ & $65.00 \pm 1.55$  & \pmb{$64.02 \pm 2.13$} \\ \bottomrule
\end{tabular}%
\vspace{5pt}
\caption{Ablation study highlighting the contribution of different design choices. We evaluate the effect of KD, domain-expert teachers, pWCTA, and logit standardization. We report IoU (in \%) for each result as mean and standard deviation. The best and second-best results are highlighted and underlined, respectively.}
\label{tab:ablation}
\end{table*}

\subsection{DG Benchmark}
We run the leave-one-out DG benchmark described in \ref{subsec:training} and report the results with their mean and standard deviation in Table \ref{tab:dg}. On average, our ensemble distillation methodology is 1.3\% better than the second-best compared solution (XDED), which also distills from a set of specialized teachers. This strategy, hence, proves to give insightful information to the student and makes it less biased towards domain-specific features. The results for ERM are quite balanced across domains, proving the strong validity of this method despite its simplicity. Other DG methods, even though obtaining superior results in specific domains, are, on average, suboptimal. This failure could be due to the methods focusing on features that are extremely beneficial for some specific scenarios but useless to others. Our method, instead, retains consistently good performance in all domains thanks to the insights given by the ensembled teachers. %However, the variance in results is considerable for the most challenging domains for almost all the DG methodologies tested. WildNet, instead, presents quite stable average performances over the runs but reports suboptimal results. This finding suggests that DG training offers a complex challenge, and our KD methodology could be further studied and improved to provide more robust results. We will address this aspect in future works.

To further validate the generalization capability of our method, we construct a more challenging benchmark by using five unseen test domains (\textit{Pear Tree}, \textit{Zucchini}, \textit{Real Vineyard}, \textit{Real Miscellaneous} and \textit{VegAnn}). The models are trained and validated on all four datasets used for the previous benchmark. %In this way, each model has been trained on at least a domain similar in shape and size to a target domain, informing the models about the principal geometric features of different plant types. 
In this test, we also investigate the Sim-to-Real gap. The results are reported in Table \ref{tab:test}, where we also include the teachers' performance as an upper bound. Teacher IoU is lower for real datasets as they are more challenging and contain fewer samples.
On average, our ensemble distillation methodology is 1.4\% better than the second-best compared solution (XDED), confirming the outcome of the previous benchmark. Moreover, our method retains the best performance on all real domains. This result enforces previous considerations on the generalization ability of KD without any additional layers or computation at inference time.
%As expected, thanks to the \textit{Generic Tree 2} source domain, all the models perform well on the \textit{Pear Tree} domain, despite its significant difference in shape from the other crops. 
ERM obtains acceptable results in all domains but is surpassed by a significant margin by other methods like IBN and ISW. %However, its performance on the \textit{Real Vineyard} domain is very low. While this is partially due to the dataset being very challenging, 
These results suggest ERM and state-of-the-art DG methods suffer Sim2Real more than ours. Indeed, the change from synthetic to real crops further widens the domain gap between different crops and backgrounds. Another interesting insight can be found in the standard deviations, as our method obtains one of the smallest values. WildNet performs badly on \textit{Real Vineyard} and \textit{VegAnn} while obtaining satisfactory results on synthetic ones. Its small standard deviation suggests that the multiple training losses applied during training could have an over-regularizing effect on the process. On the contrary, our approach finds the best trade-off between regularization and learning. In the next section, we investigate the contribution of different elements to this result. %We report a qualitative comparison of the results on real samples in Figure \ref{fig:visual}.

\subsection{Ablation Study}
\label{subsec:ablation}

We conduct an ablation study to investigate the effect of different components on the generalization capability of our methodology. We also highlight the main differences between our approach and XDED \cite{lee2022cross} regarding methodological components and performance. In particular, we consider distillation strategy, logit standardization, and pWCTA.
We also try substituting the specialized teachers with an ensemble of ERM models trained on all source domains.
The results are reported in Table \ref{tab:ablation}, in which we included ERM as a baseline. 

First, the results confirm that applying distillation improves simple ERM without requiring additional computation at inference time. Moreover, plain ensemble distillation cannot bridge the strong Sym2Real gap. This failure is probably due to the unbalanced contribution of different teachers, which can lead to transferring domain-dependent biases to the student. To contrast this risk, we standardize distillation logits and apply pWCTA with a low probability to avoid overfitting. Our methodology outperforms ERM distillation, making the best out of specialized teachers. As depicted in Fig. \ref{fig:ensemble}, the distillation masks are less confident, giving the student a better understanding of what parts of the image are more likely to confound the predictor.

\begin{figure}[t]
    \centering
    \resizebox{\columnwidth}{!}{
    \begin{tabular}{cc}
        \includegraphics[width=\columnwidth]{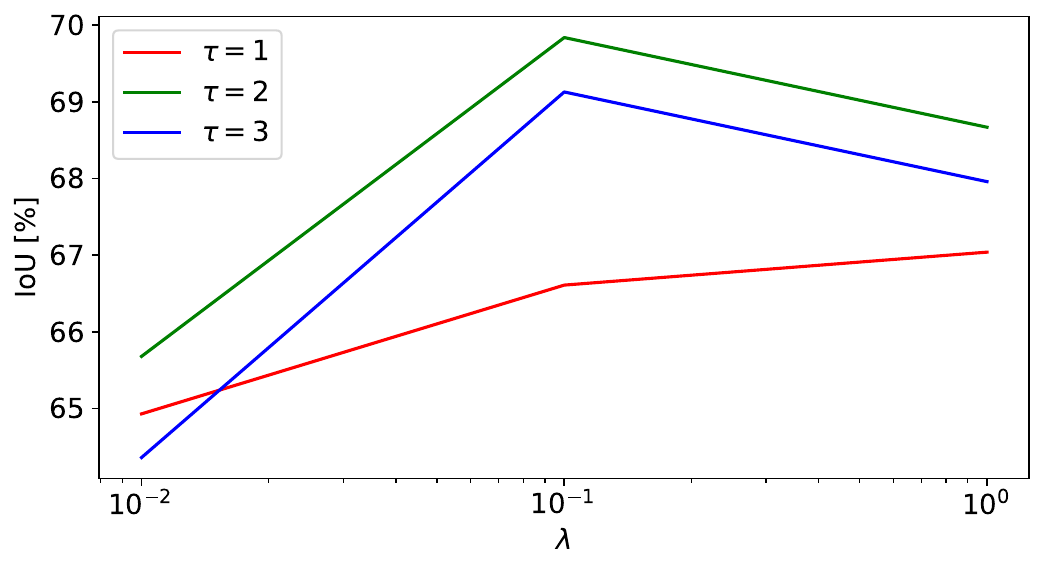} &
        \includegraphics[width=\columnwidth]{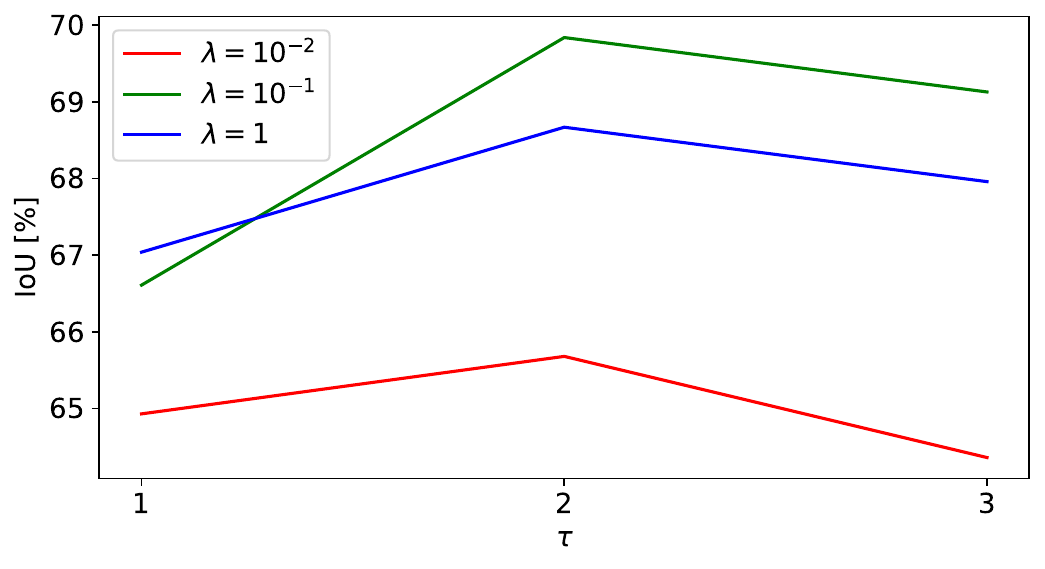}
    \end{tabular}
    }
    \caption{Ablation study on the hyper-parameters $\lambda$ and $\tau$. The reported IoU value is relative to the \textit{Real Miscellaneous} domain and is averaged on three runs. We represent two views of the results for better readability.}
    \label{fig:ablation}
\end{figure}

% Hyperparameter ablation (Lambda, T, WL)
We further inspect the effect of the method's hyper-parameters on generalization capabilities. We vary the distillation loss weight $\lambda$ and the temperature $\tau$ and report the results on the \textit{Real Miscellaneous} domain in Fig. \ref{fig:ablation}. The graphs show that our choice ($\lambda=10^{-1}, \tau=2$) is the optimal balance that ensures regularization without constraining the student. As reported in our benchmarks, this yields good generalization across various synthetic and real domains. 

\begin{comment}
%Finally, in Fig. \ref{fig:visual}, we report a qualitative comparison between output masks from our method and the most promising competitors (ERM, IBN, and XDED, according to our benchmark). We inspect output masks on \textit{Lettuce}, \textit{Real Vineyard}, and \textit{Real Miscellaneous} domains for random samples. Although IoU is computed with a confidence threshold of 90 \%, we choose to plot the original masks to highlight unconfident predictions. The difference is most evident for the \textit{Lettuce} domain, in which other algorithms erroneously segment the terrain (ERM, IBN) or retain low confidence on crops (XDED). The same happens for the \textit{Real Vineyard} domain, where the predictions are generally less confident, and XDED performs similarly to our solution. On the \textit{Real Miscellaneous} domain, XDED performs slightly worse than our solution, as the segmentation mask does not include trunks. In this scenario, IBN is more accurate and similar to our method, confirming the results of Table \ref{tab:test}. In conclusion, our solution outputs satisfactory masks for all domains, performing on par or better than all other methods.
\end{comment}

\section{Conclusions}
\label{sec:conclusions}
In this work, we proposed a novel method to tackle the problem of DG for crop semantic segmentation in realistic scenarios. We demonstrated that our distillation method represents a competitive approach for transferring domain-specific knowledge learned from multiple teacher models to a single student without any overhead at inference time. Moreover, we proposed logit standardization to adapt ensembled knowledge to the student, balancing overconfident predictions and penalizing spurious correlations. Each teacher must be trained only once, and the method can be extended to more domains by just training a new teacher and then distilling. %We conceived our solution to extend segmentation models' robustness and generalization properties to unseen environmental conditions or crops. 
Extensive experimentation has been conducted on the novel multi-crop synthetic dataset \textsc{AgriSeg} and on real test data to demonstrate the overall generalization boost provided by our training method. Moreover, we conducted an ablation study to highlight the role of different components in our solution. The superior results provided by our method show how pairing ensembled KD and DG can lead to robust perception models for realistic tasks in precision agriculture. Future works will progressively add more domains and DG methods to the \textsc{AgriSeg} benchmark. We will also include more real-world labeled data to guarantee a deeper investigation of the use of synthetic data for robust generalization in agriculture.

\section*{Acknowledgements}
 This work has been developed with the contribution of the Politecnico di Torino Interdepartmental Centre for Service Robotics (\href{https://www.pic4ser.polito.it}{PIC4SeR}). 

\clearpage

\bibliographystyle{unsrtnat}
\bibliography{bibliography.bib}  %%% Uncomment this line and comment out the ``thebibliography'' section below to use the external .bib file (using bibtex).

\end{quote}
\end{document}